\documentclass[11pt,a4paper]{article}
\usepackage{amssymb}
\usepackage{multicol}
\usepackage{graphicx}
\usepackage{subcaption}
\usepackage{indentfirst}
\usepackage{amsmath}
\usepackage{makecell}
\usepackage{amsfonts}
\usepackage{textcomp}
\usepackage{mathcomp}
\usepackage{gensymb}
\usepackage{wasysym}
\usepackage{booktabs}
\usepackage{txfonts}
\usepackage{hyperref}
\usepackage[margin=2.5cm]{geometry}
\usepackage{times}
\usepackage{mathptmx}
\usepackage{etoolbox}
\usepackage{float}
\usepackage{xurl}

\usepackage{natbib}
\usepackage{titlesec}
\usepackage[font=normalfont]{caption}

\hypersetup{pdfprintscaling=None}

\title{Traceable Virtual Sea Trials in the Marine Robotics Unity Simulator for Manoeuvring Assessment of Unmanned Surface Vehicles}
\date{}
\author{\bigskip Paria Rezayan$^{a,*}$ \\ \smallskip \textit{$^{a}$School of Engineering and Built Environment, Sheffield Hallam University, UK} \\ $^{*}$Corresponding author. Email: Paria.Rezayan@student.shu.ac.uk}

\titleformat{\section}
       {\normalfont\large\bfseries}{\thesection}{1em}{}

\titleformat{\subsection}
       {\normalfont\normalsize\bfseries}{\thesubsection}{1em}{}

\titleformat{\subsubsection}
       {\normalfont\normalsize\itshape}{\thesubsubsection}{1em}{}
       
\titlespacing\section{0pt}{12pt plus 4pt minus 2pt}{4pt plus 2pt minus 2pt}
\titlespacing\subsection{0pt}{12pt plus 4pt minus 2pt}{4pt plus 2pt minus 2pt}
\titlespacing\subsubsection{0pt}{12pt plus 4pt minus 2pt}{0pt plus 2pt minus 2pt}

\providecommand{\keywords}[1]{\hspace{4.5mm}{\small{\textit{Keywords: } #1}}}

\begin{document}

\date{}

\maketitle

\begin{center}
\begin{minipage}{0.92\textwidth}
\small
\textbf{Author Preprint Notice.}
This is the author-prepared preprint of a paper submitted to iSCSS 2026 following abstract acceptance. The manuscript is currently under review and has not yet been peer reviewed. If accepted, the published version will be available through the IMarEST Library, and this record will be updated with the official DOI.
\end{minipage}
\end{center}

\renewcommand{\abstractname}{Abstract}
\begin{abstract}
Accurate identification of hydrodynamic derivatives is essential for precise control and autonomous navigation of Unmanned Surface Vehicles (USVs). Acquiring high-fidelity manoeuvring data from physical sea trials is often constrained by cost, safety, and environmental disturbances. Standard manoeuvring trials, particularly Turning Circle (TC) and Zig-Zag (ZZ), remain fundamental to International Maritime Organization (IMO) and International Towing Tank Conference (ITTC) assessment procedures because they provide comparable performance metrics reflective of underlying hydrodynamic behaviour. 

This paper extends the open-source Marine Robotics Unity Simulator (MARUS) by introducing a standardised Virtual Sea Trial framework for automated execution and data generation of TC/ZZ manoeuvres, with traceable command–actuation logging, system-identification (SI)-focused data conditioning, and automated extraction of IMO/ITTC-aligned manoeuvring metrics and compliance criteria. A key contribution is an extensible integration within MARUS that introduces a dedicated TC/ZZ data acquisition and post-processing pipeline, thereby improving the repeatability and auditability of simulator-based manoeuvres while producing SI-ready datasets to support more reliable hydrodynamic-derivative identification and digital-twin workflows. Another distinguishing feature is explicit command–execution separation for differential‑thrust steering, where manoeuvre inputs are recorded as ordered rudder‑equivalent commands and realised actuation is logged as an execution‑level proxy derived from applied thrust. This addresses a known SI failure mode whereby commanded inputs are treated as achieved actuation (common in simulator-based SI datasets), despite actuator saturation and practical implementation constraints. 

Case‑study results demonstrate repeatable and compliant manoeuvre behaviour. For TC tests, the normalised advance differs by $\sim 3.9\%$ between port and starboard sides, while the tactical diameter differs by $\sim 4.6$--$4.7\%$, indicating good directional symmetry. For ZZ tests, on the other hand, first and second overshoot excesses remain below $1^\circ$ for both $\pm10^\circ$ and $\pm20^\circ$ manoeuvres, satisfying IMO criteria, while peak yaw rates range from approximately $4.1$ to $5.8~\mathrm{deg/s}$ across manoeuvre magnitudes.

Overall, the proposed framework provides a repeatable and auditable virtual sea-trial workflow for generating IMO/ITTC-aligned manoeuvring datasets. By improving the traceability of commanded and realised actuation, it supports more reliable simulator-based system identification, hydrodynamic-derivative estimation, and digital-twin calibration for USV autonomy. The paper outlines the MARUS implementation, SI‑oriented data conditioning and metric extraction, case‑study validation, and concludes with discussion and future work.
\end{abstract}

\keywords{Virtual Sea Trials; MARUS; Turning Circle/Zig-Zag; System Identification; IMO/ITTC; Digital Twin}

\let\thefootnote\relax\footnote{\textbf{Authors' Biographies}\\\textbf{Paria Rezayan} is a PhD Researcher in Marine Robotics at Sheffield Hallam University, working on real-time data-driven system identification for autonomous marine vessels using MARUS. She holds an MSc in Artificial Intelligence and has prior research experience in machine learning with the NHS.}

\section{Introduction}\label{sec:intro}
Covering over 70\% of the Earth’s surface, the global oceans necessitate the development of a broad spectrum of autonomous marine vehicles, especially unmanned surface vessels (USVs). USVs have been explored more intensely than their counterparts, such as unmanned underwater vehicles (UUVs), due to their higher operational safety, lower cost, lower energy consumption, and better deployment flexibility. Such advantages have positioned USVs at the forefront of maritime operations across transportation, commercial, scientific, and military domains. Therefore, accurate prediction of USV motion and manoeuvrability is critical for autonomy functions ranging from path following and course keeping to guidance, navigation, and collision avoidance \citep{Bai2022,Lv2025}.

USV motion prediction relies on physics‑based manoeuvring models which represent the hydrodynamic forces and moments acting on the hull as mathematical functions of current vessel states, control inputs, and surrounding environmental conditions. The central aim of such models, most commonly the single-system Abkowitz-type \citep{Abkowitz1964} and modular MMG-style models \citep{Yasukawa2015}, is the accurate estimation of vessel‑specific hydrodynamic derivatives, which ultimately determine how the vessel responds to control signals and external disturbances \citep{Fossen2011}. 

Traditionally, hydrodynamic derivatives were obtained from captive model tests, free-running experiments, or full-scale sea trials \citep{Xu2025}. However, these tests are costly, time-consuming, labour-intensive, and unfit for rapid prototyping or real-time applications. Consequently, System Identification (SI) techniques have gained prominence as efficient alternatives, estimating coefficients from recorded trajectories or computational simulations such as virtual sea trials. However, classical SI methods remain sensitive to noise and limited in capturing nonlinear hydrodynamic interactions, while more recent purely ML-based approaches suffer from data hunger, poor extrapolation, and a lack of interpretability \citep{Zhang2015,Umenberger2018,Alexandersson2022,Moreira2022,Luo2014,ArizaRamirez2018}. These challenges have motivated the widespread adoption of hybrid grey-box methodologies, including Physics-Informed Neural Networks (PINNs), in hydrodynamic-derivative estimation \citep{An2025}.

Reliable motion prediction requires hydrodynamic derivatives to be treated not as fixed constants but as time-dependent quantities that vary with evolving operating conditions, which necessitates online and continuous recalibration of manoeuvring models. Digital twins (DTs) provide this capability, where a continuously updated virtual counterpart of the vessel mimics its real-time behaviour through adaptive, data-driven modelling \citep{Tadros2025}. However, before high-level motion prediction, SI, or DT-based methods can be used confidently in practice, the underlying manoeuvre data must be repeatable, traceable, and physically meaningful. 

\begin{figure}[H]
    \centering

    \begin{subfigure}{0.45\linewidth}
        \centering
        \includegraphics[
            width=\linewidth,
            height=0.32\textheight,
            keepaspectratio
        ]{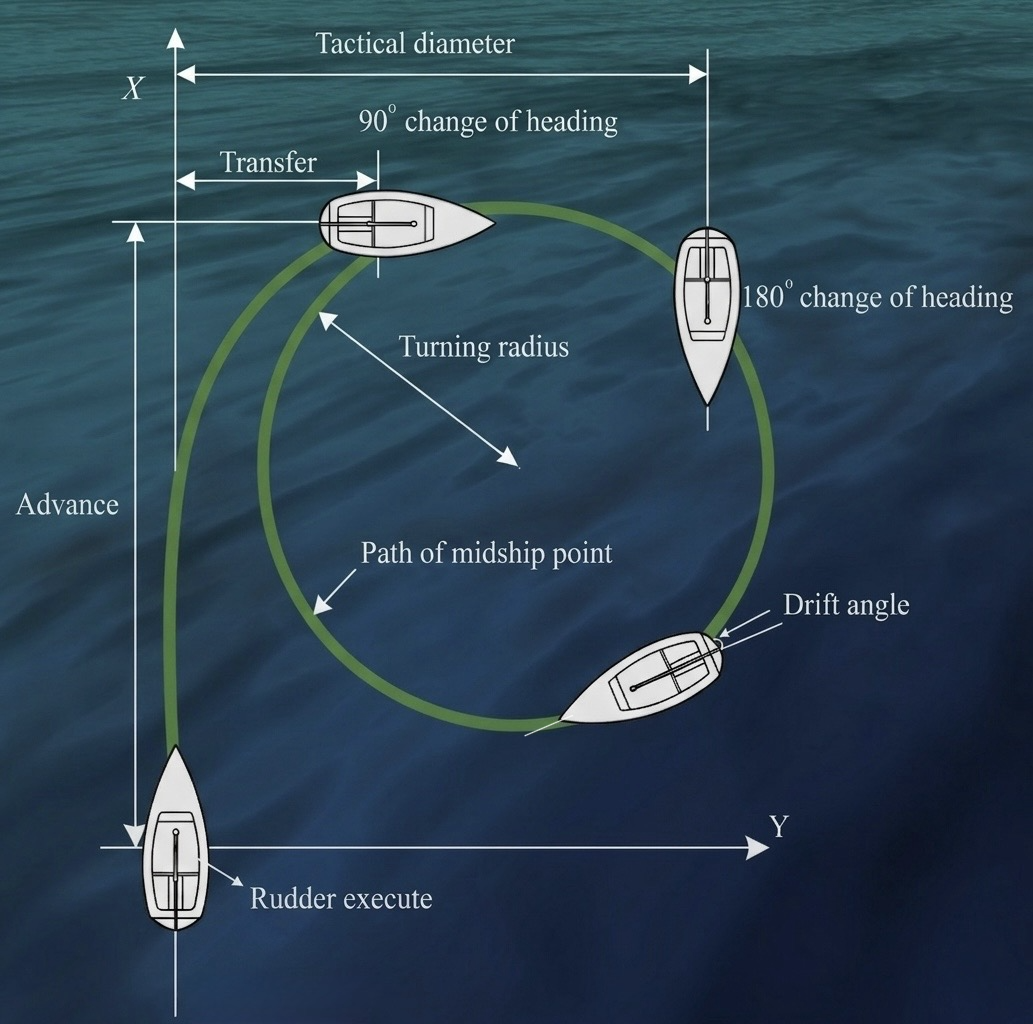}
        \caption{Turning Circle manoeuvre.}
        \label{fig:tc_metrics}
    \end{subfigure}
    \hfill
    \begin{subfigure}{0.48\linewidth}
        \centering
        \includegraphics[
            width=\linewidth,
            height=0.55\textheight,
            keepaspectratio
        ]{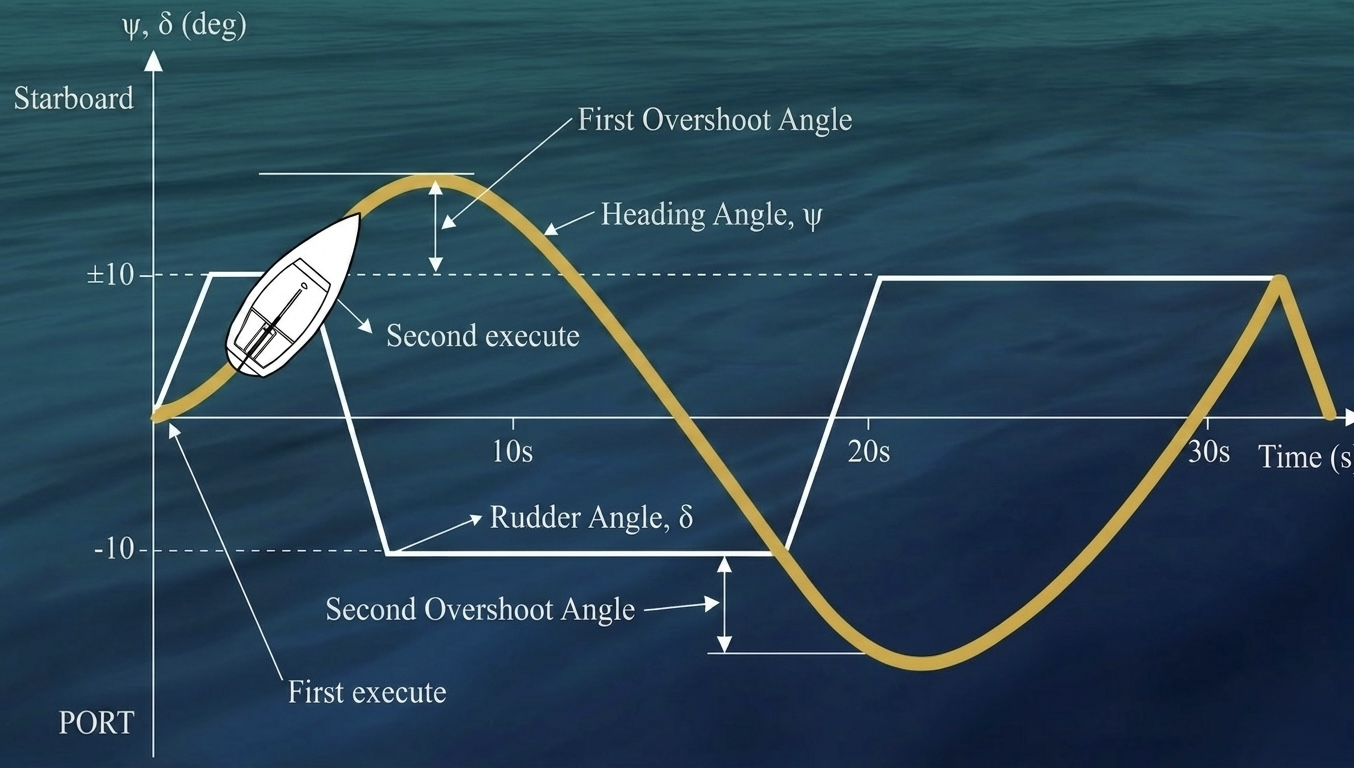}
        \caption{Zig--Zag manoeuvre.}
        \label{fig:zz_metrics}
    \end{subfigure}

    \caption{Standard USV manoeuvres and metrics: (a) TC with advance, transfer, and tactical diameter; (b) ZZ with heading crossings and overshoot angles.}
    \label{fig:standard_manoeuvre_metrics}
\end{figure}

To establish this foundation, standard manoeuvring tests such as the Turning Circle (TC) and Zig-Zag (ZZ) manoeuvres are performed to evaluate vessel dynamic behaviour through structured excitation and response. Such tests assess key aspects of USV manoeuvrability, including straight‑line stability, turning ability, yaw‑checking ability, and stopping ability. This is where the IMO and ITTC procedures become essential: the IMO Standards for Ship Manoeuvrability define performance metrics such as advance, tactical diameter, and overshoot angles, while ITTC trial procedures specify relevant trial conditions including approach conditions and data measurement and acceptance criteria. Figure~\ref{fig:standard_manoeuvre_metrics} shows the TC and ZZ metric definitions. Although originally developed for conventional ships, these frameworks remain directly applicable to USVs because they provide a common basis for manoeuvring assessment, hydrodynamic‑derivative extraction, and validation of SI- and DT-based modelling frameworks \citep{IMO2002,ITTC2017}.

However, performing those standard manoeuvring tests through physical sea trials is costly and weather-dependent, particularly for small-scale USVs where limited low-cost sensors, actuator constraints, and environmental disturbances increase signal variability. Furthermore, SI studies have shown that hydrodynamic-parameter estimates are highly sensitive to manoeuvre design, input-excitation richness, noise levels, and the consistency of measured control signals \citep{Xu2025,Wang2025}.

High-fidelity computational fluid dynamics (CFD) platforms, such as OpenFOAM or STAR-CCM+, can capture complex hydrodynamic interactions with high accuracy; however, they are computationally demanding and therefore unsuitable for real-time autonomy experimentation. Such limitations create a critical need for virtual sea-trial environments, which enable controlled, repeatable, and traceable execution of standard manoeuvres while offering sufficient physical fidelity for manoeuvring analysis, SI, and DT calibration \citep{Tadros2025}.

Marine robotics simulators offer a practical pathway toward realising virtual sea‑trial environments because they allow controlled, rapid, and traceable testing of vessel hydrodynamic behaviour before real-world deployment. Hence, a wide range of simulation platforms has been developed to support USV autonomy research and development, each offering different trade-offs between physical fidelity, computational efficiency, and real-time capability. In contrast to CFD-driven tools, MATLAB-based frameworks such as MSS and MANSIM \citep{FossenPerez2004,Perez2006,Sukas2019} offer fast and physically interpretable 3–6 DOF manoeuvring models that are useful for early controller or state estimator development. However, they lack modern middleware integration (such as ROS2) and cannot update parameters during real-time execution within closed-loop autonomy or DT pipelines.

Ultimately, robotics-oriented simulators represent the most practical class of tools for real-time autonomy experimentation. Robot Operating System (ROS)-based environments, including Gazebo and the UUV Simulator \citep{Manhaes2016}, as well as frameworks such as DAVE \citep{Zhang2022DAVE}, MOOS-IvP \citep{Benjamin2013,Newman2024MOOS}, and ASVSim \citep{Smith2019,Lesy2025}, support real-time execution, high-fidelity sensor emulation, and synchronised middleware-driven communication. However, they typically provide limited visual realism and simplified hydrodynamics, restricting their usefulness for generating high-quality manoeuvring datasets.

\begin{figure}[H]
    \centering
    \includegraphics[width=0.65\textwidth]{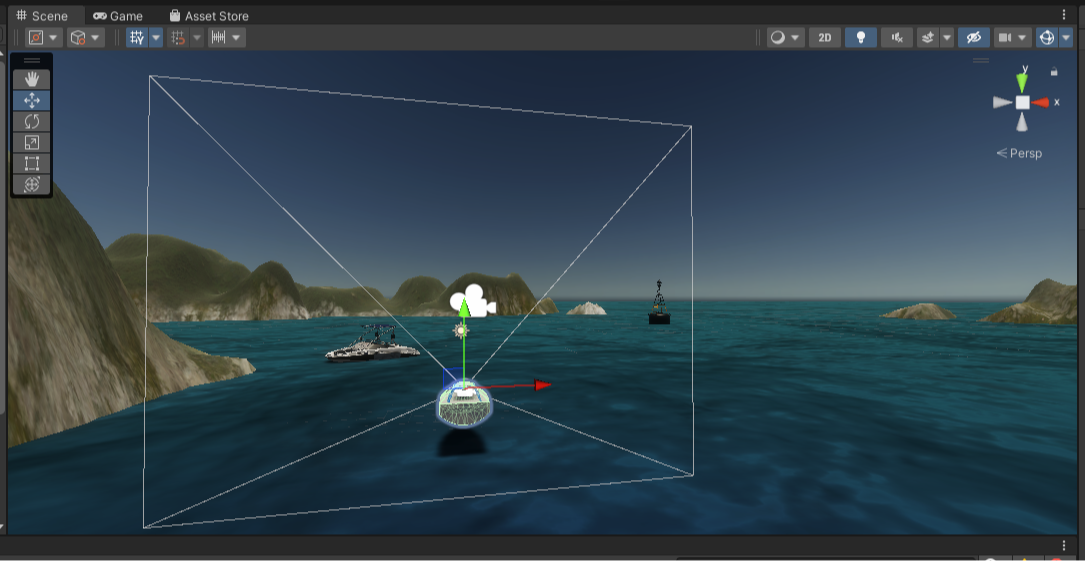}
    \caption{Example MARUS simulation scene showcasing the USV in the virtual environment.}
    \label{fig:marus_examplescene}
\end{figure}

More recent simulation frameworks based on game engines, particularly Unity and Unreal, aim to bridge this gap by combining real-time performance with improved visual fidelity and advanced sensor simulation. Platforms such as MARUS \citep{Loncar2022MARUS}, UNav-Sim \citep{Amer2023UNavSim}, and other marine simulators including Stonefish \citep{Grimaldi2025Stonefish} and HoloOcean \citep{Potokar2022HoloOcean} extend robotics-oriented frameworks by integrating photorealistic rendering and native middleware connectivity. Among these, the open-source Marine Robotics Unity Simulator (MARUS) stands out due to its balanced integration of hyper-realistic visuals, high-fidelity USV motion dynamics, rich sensor stacks, seamless ROS connectivity, and real-time execution capabilities, making it exceptionally suitable for autonomy‑oriented virtual sea trials (Figure~\ref{fig:marus_examplescene}). 

\subsection{Motivation and Identified Research Gap}

Despite these advancements, a limitation remains across robotics-oriented simulators: while they support repeatable virtual trials, they lack mechanisms for generating traceable, SI-ready manoeuvring datasets. This is critical for hydrodynamic-parameter estimation and DT calibration, where consistency between commanded inputs, realised actuation, and vessel response must be explicitly preserved. This issue is particularly important for differential-thrust USVs, where a physical rudder angle does not exist and must instead be represented through a rudder-equivalent proxy derived from the port–starboard thrust imbalance. If commanded inputs are treated as achieved actuation without accounting for delays, execution constraints, and actuator saturation, subsequent manoeuvring metrics and hydrodynamic-derivative estimates can become biased and unreliable.

This paper addresses this gap by presenting an actuation‑traceable virtual sea‑trial framework for automated TC and ZZ manoeuvres within MARUS. Unlike most simulator‑based manoeuvring studies that assume equivalence between commanded and realised actuation, the proposed framework explicitly separates, logs, and validates command–execution mismatch at the propulsion level. Additionally, differential‑thrust steering is reformulated into a realised rudder‑equivalent actuation proxy, enabling IMO/ITTC manoeuvring analysis while preserving traceability to the underlying thrust imbalance and yaw‑moment generation. Consequently, manoeuvre timing and execution are grounded in realised actuation rather than controller intent, ensuring physically consistent timing and metric extraction. 

\subsection{Major Contributions}
The contributions of this paper are threefold:
\begin{itemize}
    \item[(i)] an automated MARUS‑based framework for conducting standard virtual TC and ZZ sea trials for USV manoeuvring assessment;
    \item[(ii)] an actuation‑traceable command–execution logging structure for differential‑thrust USVs, enabling reliable SI and DT development; and
    \item[(iii)] an IMO/ITTC‑aligned post‑processing workflow for audited, SI‑focused datasets for hydrodynamic-derivative estimation and DT calibration.
\end{itemize}

\section{Virtual Sea Trial Framework}
The proposed virtual sea trial framework is implemented in MARUS, a marine simulator that integrates ROS with the Unity game engine via gRPC-based communication and Protocol Buffers (Protobuf), enabling real-time, bi-directional data exchange. The simulator combines high-fidelity 3D rendering, a wide range of virtual sensors (e.g., IMU, GNSS, LiDAR, and sonar), and highly realistic physics-based modelling of buoyancy, hydrodynamics, and environmental forces to realistically reproduce closed-loop USV manoeuvres.

\subsection{System Architecture and Actuation Execution} 
\begin{figure}[H]
    \centering
    \includegraphics[width=0.7\linewidth,height=0.30\textheight,keepaspectratio]{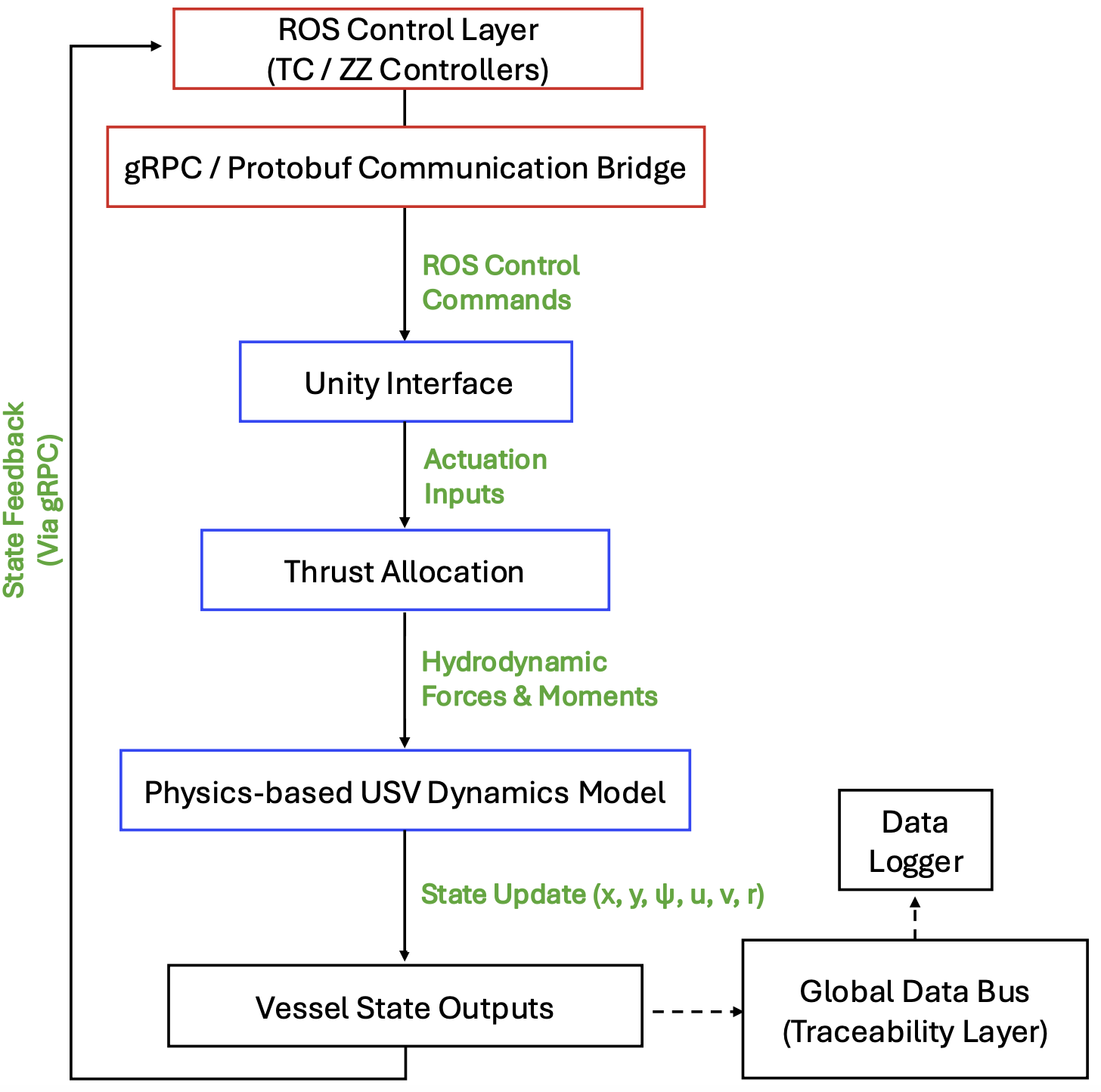}
    \caption{ROS--Unity architecture for actuation-traceable virtual sea trials. Red: ROS control layer; blue: Unity simulation and physics engine.}
    \label{fig:backend_architecture}
\end{figure}

The framework follows a hybrid control architecture (Figure~\ref{fig:backend_architecture}) that integrates Unity‑based vessel dynamics with an ROS‑based manoeuvre control layer. In this architecture, Unity performs physics integration, hydrodynamic response, propulsion‑level actuation, and pre‑manoeuvre speed stabilisation, while ROS generates high‑level command sequences for TC and ZZ tests. Sensor streams, propulsion commands, and vessel states are exchanged via ROS topics and logged through a global data bus for the subsequent SI-focused post‑processing stage.

\begin{figure}[H]
    \centering
    \includegraphics[
        width=0.5\linewidth,
        height=0.21\textheight,
        keepaspectratio
    ]{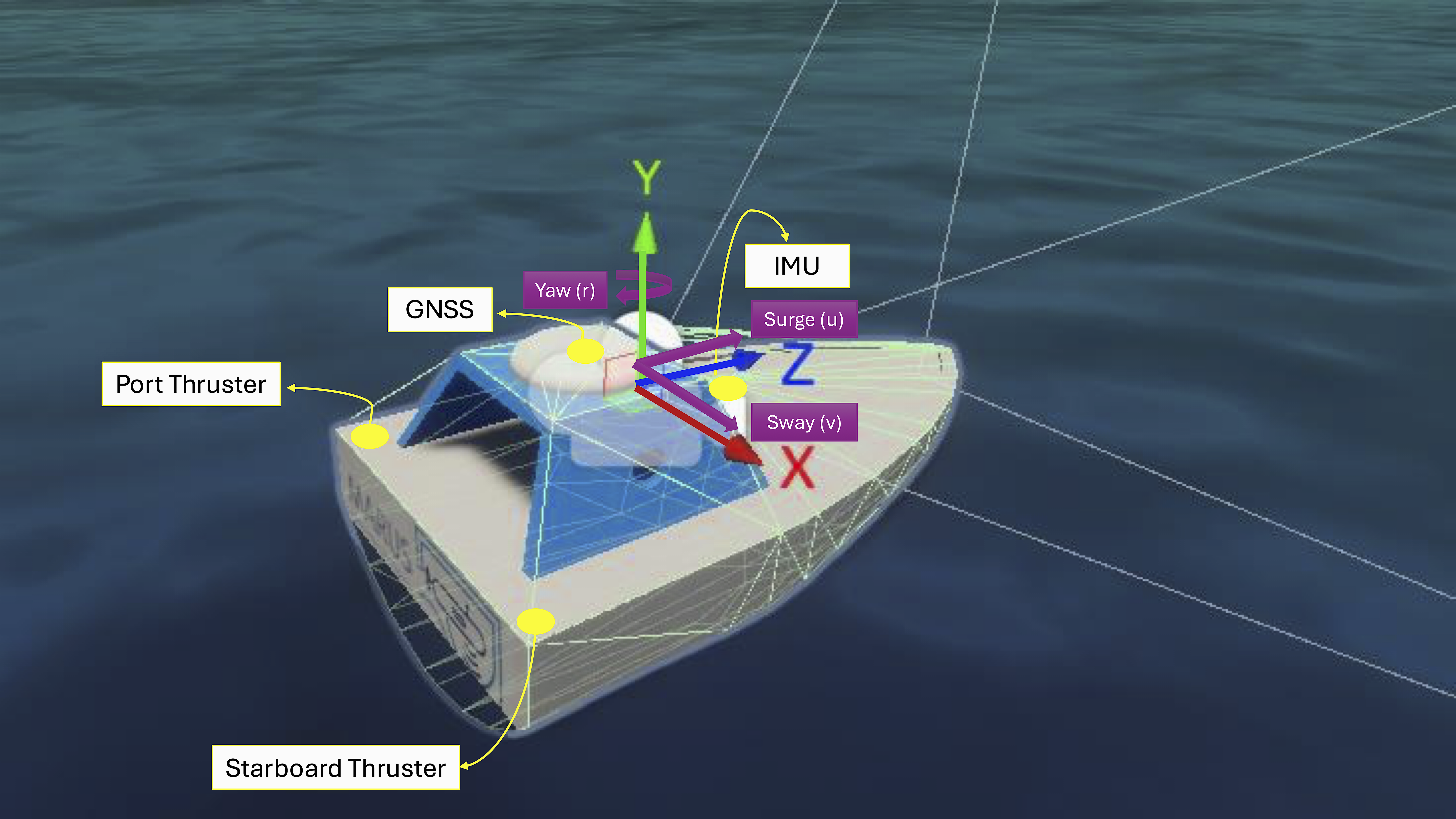}
    \caption{Simulated USV with differential-thrust actuators, onboard sensors, and body-fixed surge--sway--yaw reference frame.}
    \label{fig:marus_boat}
\end{figure}

 The simulated vessel (Table~\ref{tab:marus_specs}) is modelled as a three-degree-of-freedom rigid body in surge, sway, and yaw. At the actuation level, thruster commands are converted to realised forces using calibrated nonlinear thrust curves and applied at fixed port and starboard hull locations (Figure~\ref{fig:marus_boat}), such that differential thrust generates the yaw response, $\tau_z \propto (T_{\text{stbd}} - T_{\text{port}})$.

\begin{table}[H]
\centering
\caption{Particulars of the USV used in the virtual sea trials, where $L_{PP}$ denotes the length between perpendiculars.}
\label{tab:marus_specs}
\small
\begin{tabular}{lcccc}
\toprule
 & \textbf{Length} & \textbf{Beam} & \textbf{Draft} & \textbf{Mass} \\
 & ($\approx L_{PP}$) [m] & [m] & [m] & [kg] \\
\midrule
MARUS USV & 2.952 & 1.468 & 1.361 & 300 \\
\bottomrule
\end{tabular}
\end{table}

\subsection{Rudder-Equivalent Actuation and Execution-Triggered Timing} 

Differential-thrust USVs do not possess a physical rudder, whereas standard IMO and ITTC manoeuvring procedures are defined in terms of rudder angle and rudder-induced yaw response. To allow direct application of these procedures while preserving physical interpretability, the differential-thrust configuration is reformulated into a rudder-equivalent representation. As a result, two signals are defined: \begin{align} \delta_{\text{cmd}} &= \text{ordered rudder-equivalent command} \\ \delta_{\text{exec}} &= \text{executed rudder-equivalent proxy} \end{align}

The executed proxy is computed and logged directly from thruster inputs:
\begin{equation}
\delta_{\text{exec}} = \delta_{\max} \cdot 
\frac{T_{\text{stbd}} - T_{\text{port}}}{T_{\text{stbd}} + T_{\text{port}}}
\end{equation}

where $\delta_{\max}$ denotes the maximum rudder-equivalent angle set by the manoeuvre specification (e.g., $35^\circ$ for TC and $10^\circ$ or $20^\circ$ for ZZ tests). This formulation ensures $\delta_{\text{exec}}$ reflects the actual realised actuation rather than the commanded input.

\begin{figure}[H]
    \centering
    \includegraphics[width=0.35\linewidth]{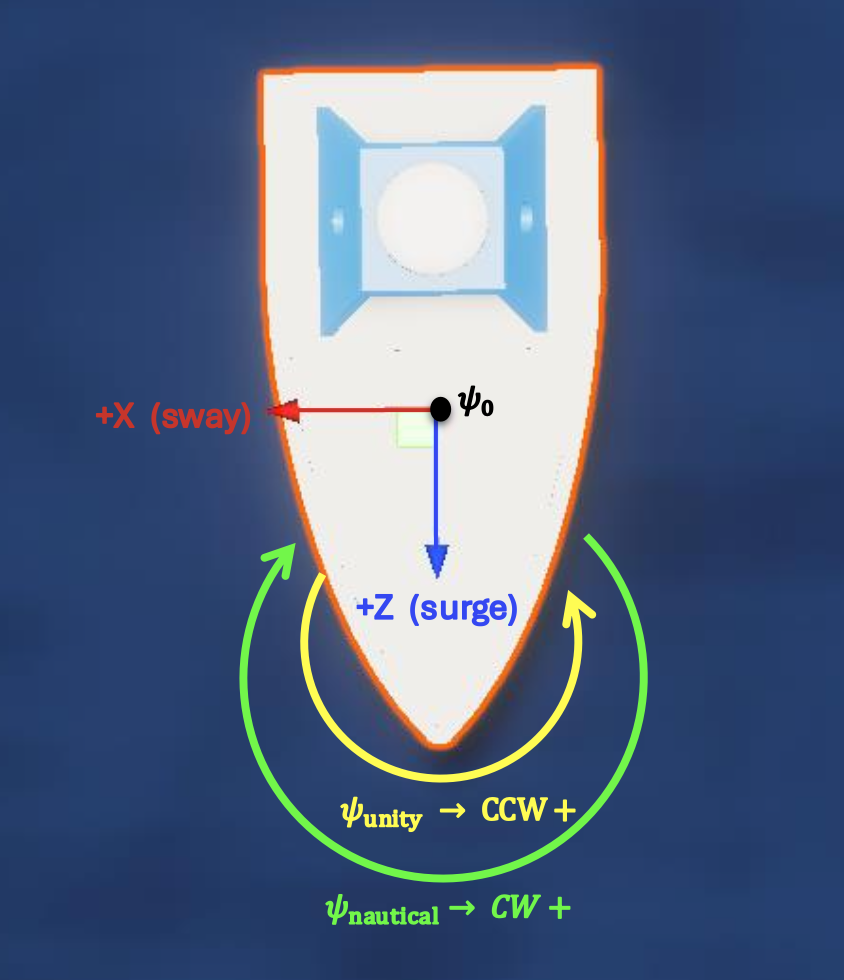}
    \caption{Coordinate systems and heading conventions used for manoeuvre analysis.}
    \label{fig:coord_system}
\end{figure}

All state measurements are obtained using the Unity world frame, where surge, sway, and yaw correspond to the $Z$, $X$, and $Y$ axes, respectively (Figure~\ref{fig:coord_system}). Since the native heading $\psi_{\text{unity}}$ follows a counter-clockwise-positive convention, it is transformed into the nautical convention used in manoeuvring analysis: 
\begin{equation} \psi_{\text{nautical}} = -\left(\psi_{\text{unity}} - \psi_0\right) \end{equation} where $\psi_0$ is the heading at manoeuvre onset. 

Manoeuvre timing, on the other hand, is defined using executed actuation. The start time $t_0$ and reference heading $\psi_0$ are given by \begin{equation}
t_0 = \min \{ t : |\delta_{\text{exec}}(t)| > \epsilon \}, 
\qquad \psi_0 = \psi(t_0),
\end{equation}

where $\epsilon$ is a small actuation threshold. This methodological refinement aligns all measurements with the true onset of actuation, avoiding errors caused by control delay or actuator saturation.
 
\section{Turning Circle and Zig-Zag Manoeuvre Execution}
Standard manoeuvring trials are implemented to evaluate vessel response under controlled steering inputs and extract IMO/ITTC-aligned performance metrics. Within the proposed framework, manoeuvre execution is governed by ROS-based controllers operating on real-time feedback of vessel heading and realised actuation. Both TC and ZZ manoeuvres follow a structured sequence consisting of stabilisation, actuation triggering, response development, and termination.

\subsection{Turning Circle Manoeuvre}
The TC manoeuvre assesses steady turning behaviour through the metrics of advance, transfer, and tactical diameter. A constant rudder-equivalent command of $\delta_{\text{cmd}} = \pm 35^\circ$ is applied via differential thrust. The execution process is defined as follows:

\textbf{Initial Condition Stabilisation:}  
Prior to manoeuvre initiation, the vessel is initiated by a constant forward speed using a surge speed controller (SpeedHold), while yaw stability is monitored using onboard inertial measurements. The manoeuvre is initiated only once low yaw rate and minimal deviations in speed and heading are achieved. 

All TC and ZZ manoeuvres use a target approach speed of approximately 90\% of the steady speed at 85\% Maximum Continuous Rating (MCR). Calibration of the MARUS USV resulted in $V_{85} = 0.9235~\mathrm{m/s}$ and $V_{\text{target}} = 0.8311~\mathrm{m/s}$.

\textbf{Manoeuvre Initiation and Data Acquisition:}  
Following stabilisation, a constant differential-thrust input of $\delta_{\text{cmd}} = \pm 35^\circ$ is applied, and vessel states and actuation signals are logged simultaneously.

\textbf{Manoeuvre Execution and Termination:}  
Differential thrust is maintained to generate a sustained turning motion. The manoeuvre is terminated once the cumulative heading change satisfies
\begin{equation}
|\psi_{\text{nautical}}| \geq 540^\circ,
\end{equation}
ensuring both transient and steady-state turning behaviour are captured.

From the resulting trajectory, standard manoeuvring metrics are extracted (Figure~\ref{fig:tc_metrics}):
\begin{equation}
\text{Advance} = X(90^\circ), \quad
\text{Transfer} = Y(90^\circ), \quad
\text{Tactical Diameter} = Y(180^\circ).
\end{equation}
These metrics quantify longitudinal progression, lateral displacement, and overall turning trajectory extent, respectively. All quantities are normalised by $L_{PP}$ for dimensionless port--starboard comparison.

\subsection{Zig-Zag Manoeuvre}

The ZZ manoeuvre evaluates yaw-checking ability, transient response, and overshoot behaviour. Rudder-equivalent steering commands are applied via differential thrust for both $\pm 10^\circ / \pm 10^\circ$ and $\pm 20^\circ / \pm 20^\circ$ tests.

\textbf{Initial Condition Stabilisation:}  
As with the TC, execution begins from a stabilised forward speed.

\textbf{Manoeuvre Initiation and Data Acquisition:}  
Once steady-state conditions are achieved, rudder-equivalent commands are applied via a state-driven switching logic based on measured heading response. The steering direction is reversed when the heading reaches the prescribed deviation threshold:
\begin{equation}
\psi_{\text{nautical}} = \pm \Delta \quad \Rightarrow \quad \delta_{\text{cmd}} \rightarrow -\delta_{\text{cmd}},
\end{equation}
where $\Delta \in \{10^\circ, 20^\circ\}$. This formulation ensures that command reversals are triggered by the vessel response rather than predefined timing.

\textbf{Manoeuvre Execution and Termination:}  
Manoeuvre execution is repeated to generate alternating heading responses and terminated after the required number of crossings is achieved. From the recorded data, key performance metrics are extracted (Figure~\ref{fig:zz_metrics}):

\begin{itemize}
\item \textbf{First Overshoot Angle ($\alpha_1$):} heading deviation exceeding the first steering reversal,
\item \textbf{Second Overshoot Angle ($\alpha_2$):} heading deviation following the second reversal,
\item \textbf{Execution Events:} timing of command reversals based on realised actuation.
\end{itemize}

These metrics capture yaw stability, rudder responsiveness, and execution consistency.

\subsection{Time-Aligned Logging and SI-Ready Traceability}

To ensure auditable manoeuvre datasets, a global data bus (\textit{ControlBus}) synchronises actuation signals and telemetry across all system modules. The recorded signals include thruster inputs $(T_{\text{port}}, T_{\text{stbd}})$, differential and total thrust, rudder-equivalent command and execution signals $(\delta_{\text{cmd}}, \delta_{\text{exec}})$, a yaw-moment proxy $\tau_z$, and propulsion telemetry.

In parallel, a Unity-facing logger (\textit{BoatDataLogger}) records a time-aligned dataset of vessel states, actuation signals, and derived quantities,
\begin{equation}
\mathbf{y}(t) = \{t, x, y, \psi_{\text{wrapped}}, \psi_{\text{unwrapped}}, \psi_{\text{relative}}, \psi_{\text{nautical}}, u, v, r, \delta_{\text{cmd}}, \delta_{\text{exec}}, T_{\text{port}}, T_{\text{stbd}}, \tau_z, \ldots \},
\end{equation}
where $t$ denotes simulation time and $\psi_{\text{relative}}$ is referenced to manoeuvre onset. To preserve heading continuity, the wrapped heading is unwrapped as
\begin{equation}
\psi_{\text{unwrapped},k} = \psi_{\text{unwrapped},k-1} + \Delta \psi_k.
\end{equation}

Moreover, yaw rate is obtained both directly from the Unity rigid-body model and by differentiating the processed nautical heading,
\begin{align}
r_{\text{logged}} &= \dot{\psi}_{\text{unity}}, \\
r_{\text{calc}}   &= \frac{d \psi_{\text{nautical}}}{dt}.
\end{align}

Agreement between these signals provides an internal consistency check. 

By preserving commanded input, executed actuation, and vessel response on a common time base, the framework produces physically consistent datasets for hydrodynamic-derivative estimation and DT calibration. This structure addresses a prominent limitation in simulator-based SI datasets, where commanded inputs are often incorrectly treated as achieved actuation.

\section{Data Post-Processing and Metric Extraction}
Following manoeuvre execution, the recorded datasets (Turning Circle port and starboard; Zig-Zag $10^\circ/10^\circ$ and $20^\circ/20^\circ$ in both directions) are analysed and conditioned via an SI-focused post-processing framework. Rather than assuming a seamless match between commanded input and realised actuation, the pipeline applies an actuation-grounded and traceability-aware procedure for manoeuvre alignment, metric extraction, and dataset assembly.

\subsection{Signal Conditioning and Execution-Grounded Alignment}
Post‑processing uses the execution‑triggered manoeuvre onset time $t_0$ as the shared reference for each dataset. Similarly, heading‑based quantities are evaluated relative to the executed reference heading $\psi_0$, while kinematic states and actuation signals are examined within manoeuvre‑specific time windows bounded by $t_0$ and the detected end of actuation. This structured alignment provides a coherent basis for manoeuvre segmentation and subsequent metric extraction.

\subsection{Manoeuvre Segmentation and Metric Extraction}
To ensure valid initial conditions, approach conditions are first verified over a pre‑manoeuvre time window preceding $t_0$ ($T_w$). The approach speed is defined as the time‑averaged resultant velocity
\begin{equation}
U_0 = \frac{1}{T_w} \int_{t_0 - T_w}^{t_0} V(t)\, dt,
\qquad
V(t) = \sqrt{u(t)^2 + v(t)^2},
\end{equation}
while yaw stability is assessed using the directly logged yaw‑rate signal rather than derivative‑based estimates, preventing contamination across the manoeuvre onset at $t_0$. This verification step ensures that manoeuvres are initiated under steady‑state conditions consistent with ITTC recommendations.

Following validation of the initial conditions, metric extraction is performed directly from the vessel response. For TC manoeuvres, the trajectory is first transformed into a local vessel‑fixed coordinate frame aligned with the heading at manoeuvre onset. This transformation enables separation of surge and sway motion, allowing standard IMO metrics to be evaluated in a physically interpretable frame. Advance, transfer, and tactical diameter are extracted using linearly interpolated heading crossings at $90^\circ$ and $180^\circ$ in the re‑referenced heading data. To capture steady‑turning behaviour more robustly, near‑termination trajectory segments are additionally used for guarded circle fitting when sufficient data are available.

For ZZ manoeuvres, segmentation is driven by the realised steering response instead of prescribed timing. Rudder reversals are detected from sign changes in the executed rudder‑equivalent signal, ensuring that identified reversal events correspond to physically applied actuation. Next, overshoot angles $\alpha_1$ and $\alpha_2$ are computed as extrema of the re‑zeroed heading signal within successive reversal intervals. Additionally, yaw rate is obtained from numerical differentiation of the processed heading signal, maintaining consistency with the adopted reference frame and avoiding convention‑related inconsistencies.

\subsection{Internal Consistency, Traceability, and SI‑Ready Dataset Assembly}
The final stage verifies internal consistency and assembles SI-ready datasets. The yaw rate logged directly from the Unity rigid‑body model,
$r_{\text{logged}}$, is compared with the reconstructed yaw rate, $r_{\text{calc}} = d\psi_{\text{rel}}/dt$. Agreement between these independently derived signals provides a system‑level diagnostic of simulator consistency and confirms that derived quantities remain physically compatible with the underlying
physics model.

In addition to state‑level validation, to quantify command-execution traceability, the actuation-realisation ratio
\begin{equation}
\eta = \frac{|\delta_{\text{exec}}|}{|\delta_{\text{cmd}}|}
\end{equation}
is evaluated as an indicator of actuation realisation during manoeuvre execution. 

Additional diagnostics assess time‑step consistency, actuator duty cycle, propulsion signal continuity, and command–execution consistency. Finally, processed data are organised into SI-ready datasets referenced to the time origin $t - t_0$, containing vessel states $(x, y, u, v, r)$, dual actuation signals $(\delta_{\text{cmd}}, \delta_{\text{exec}})$, and derived quantities including $r_{\text{calc}}$ and $\tau_z$, and associated diagnostics metadata. 

Throughout post‑processing, primary signals are preserved without filtering that would compromise the actuation–response relationship, with only minimal smoothing applied for stable numerical differentiation. By aligning all derived quantities with their execution‑triggered references, the framework establishes a reproducible and physically consistent foundation for simulator‑based manoeuvring analysis, SI, and DT development.

\section{Results and Validation}

This section presents the results obtained from automated TC and ZZ virtual sea trials conducted using the proposed actuation‑traceable framework. The results validate manoeuvre execution, metric extraction and expected IMO/ITTC behavioural characteristics, internal signal consistency, and command–execution traceability. Overall, they demonstrate repeatable manoeuvre behaviour and physically consistent datasets suitable for downstream tasks in hydrodynamic modelling.

\subsection{Approach Conditions, Manoeuvre Initialisation, and Kinematic Validation}
Prior to analysing manoeuvring metrics, approach conditions and manoeuvre initialisation were evaluated against ITTC steady‑state requirements. Across all trials, stable and repeatable approach conditions were achieved. Mean approach speed ranged from $0.8258$ to $0.8309~\mathrm{m/s}$, closely matching the target speed of $0.8311~\mathrm{m/s}$, with standard deviations below $0.0073~\mathrm{m/s}$ (Table~\ref{tab:approach_conditions}). 

Yaw stability was also satisfied, with the maximum absolute yaw rate in the final $10~\mathrm{s}$ preceding manoeuvre initiation remaining below
\begin{equation}
\max |r| \leq 0.179~\mathrm{deg/s}.
\end{equation}
Execution-grounded actuation thresholds detected consistent start times of approximately $t_0=121$--$124~\mathrm{s}$, which additionally confirms the following kinematic responses and manoeuvring metrics are accurately tied to physically realised actuation rather than controller timing.

\begin{table}[H]
\centering
\caption{Approach condition validation across all manoeuvres. Acceptance criteria follow ITTC steady‑state approach recommendations
($U_0 \ge 0.9U_{\text{target}}$, $\max|r| \le 0.18~\mathrm{deg/s}$).}
\label{tab:approach_conditions}
\small
\begin{tabular}{lcccc}
\hline
\textbf{Manoeuvre} 
& \textbf{Mean Approach Speed} 
& \textbf{Speed Std. Dev.} 
& \textbf{Max. Yaw Rate} 
& \textbf{Result} \\
& $U_0$ [m/s] 
& $\sigma_U$ [m/s] 
& $\max|r|$ [deg/s] 
&  \\
\hline
Turning Circle to Port               
& 0.8265 & 0.0023 & 0.178 & Pass \\
Turning Circle to Starboard          
& 0.8270 & 0.0024 & 0.178 & Pass \\
Zig--Zag $10^\circ/10^\circ$ (Port‑first)    
& 0.8258 & 0.0030 & 0.179 & Pass \\
Zig--Zag $10^\circ/10^\circ$ (Starboard‑first) 
& 0.8309 & 0.0073 & 0.179 & Pass \\
Zig--Zag $20^\circ/20^\circ$ (Port‑first)    
& 0.8260 & 0.0023 & 0.178 & Pass \\
Zig--Zag $20^\circ/20^\circ$ (Starboard‑first) 
& 0.8291 & 0.0071 & 0.179 & Pass \\
\hline
\end{tabular}
\end{table}

\begin{figure}[H]
    \centering
    \begin{subfigure}{1\linewidth}
        \centering
        \includegraphics[
            width=\linewidth,
            height=0.16\textheight,
            keepaspectratio
        ]{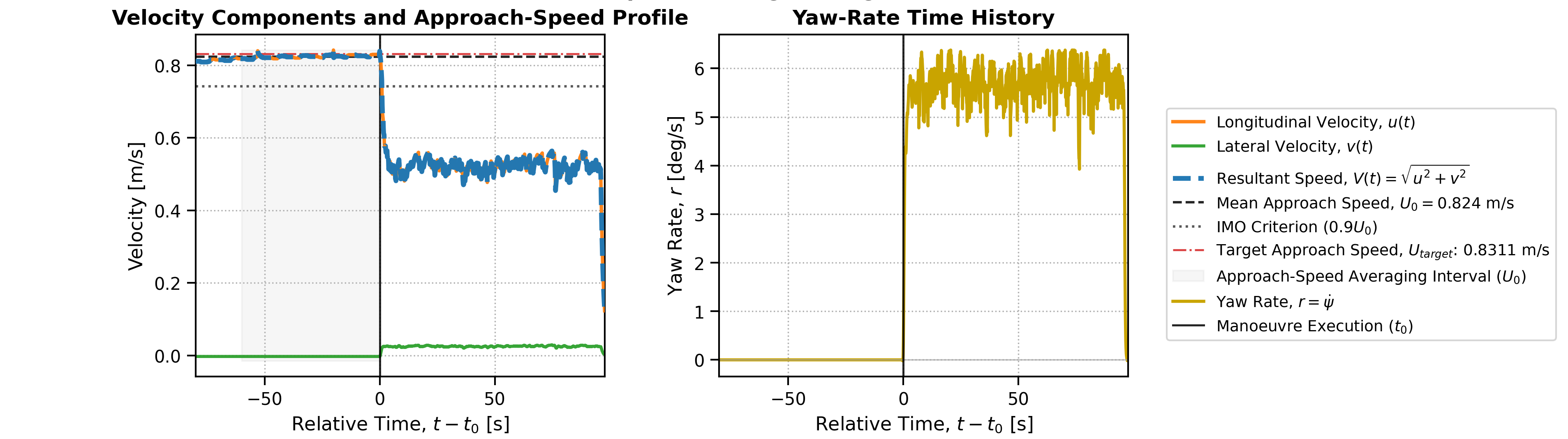}
        \caption{Turning Circle manoeuvre.}
        \label{fig:kinematics_tc}
    \end{subfigure}

    \vspace{0.5em}

    \begin{subfigure}{1\linewidth}
        \centering
        \includegraphics[
            width=\linewidth,
            height=0.16\textheight,
            keepaspectratio
        ]{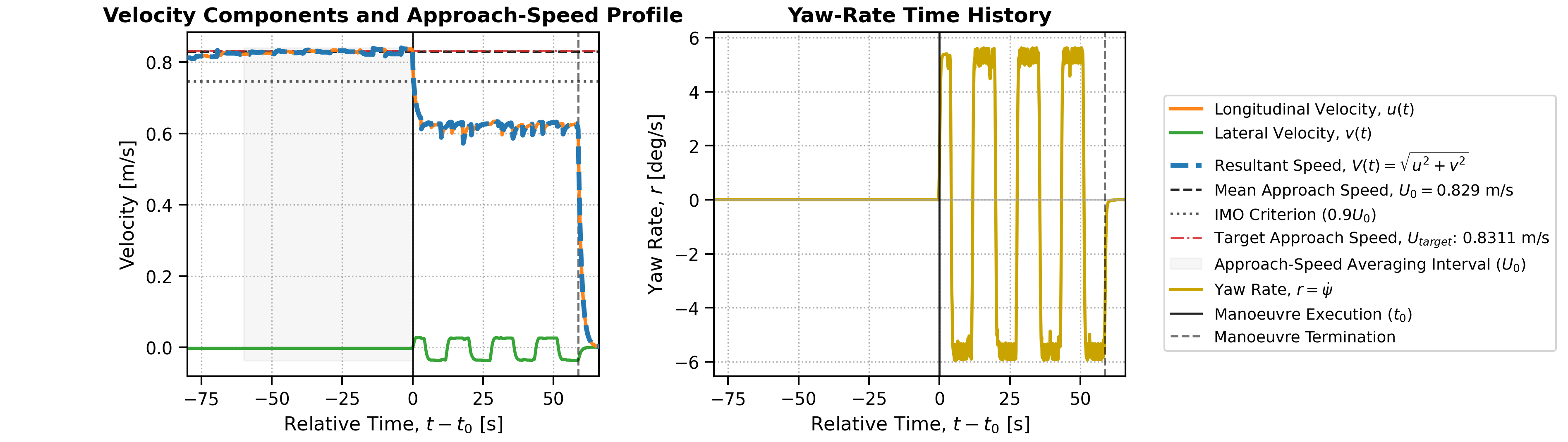}
        \caption{Zig--Zag 20/20 manoeuvre.}
        \label{fig:kinematics_zz}
    \end{subfigure}

    \caption{Representative kinematic profiles for TC to Starboard and ZZ $20^\circ/20^\circ$ Starboard-first manoeuvres.}
    \label{fig:kinematics}
\end{figure}

Figure~\ref{fig:kinematics} presents representative TC and ZZ kinematic profiles. The velocity components indicate steady approach behaviour followed by a clear transition at $t_0$. The rapid onset of yaw-rate response and the accompanying decrease in surge velocity with increasing sway velocity depict the expected kinematic development of turning motion and support subsequent metric extraction.

\subsection{Turning Circle Manoeuvre Results}

\begin{figure}[H]
    \centering
    \includegraphics[
        width=\linewidth,
        height=0.27\textheight,
        keepaspectratio
    ]{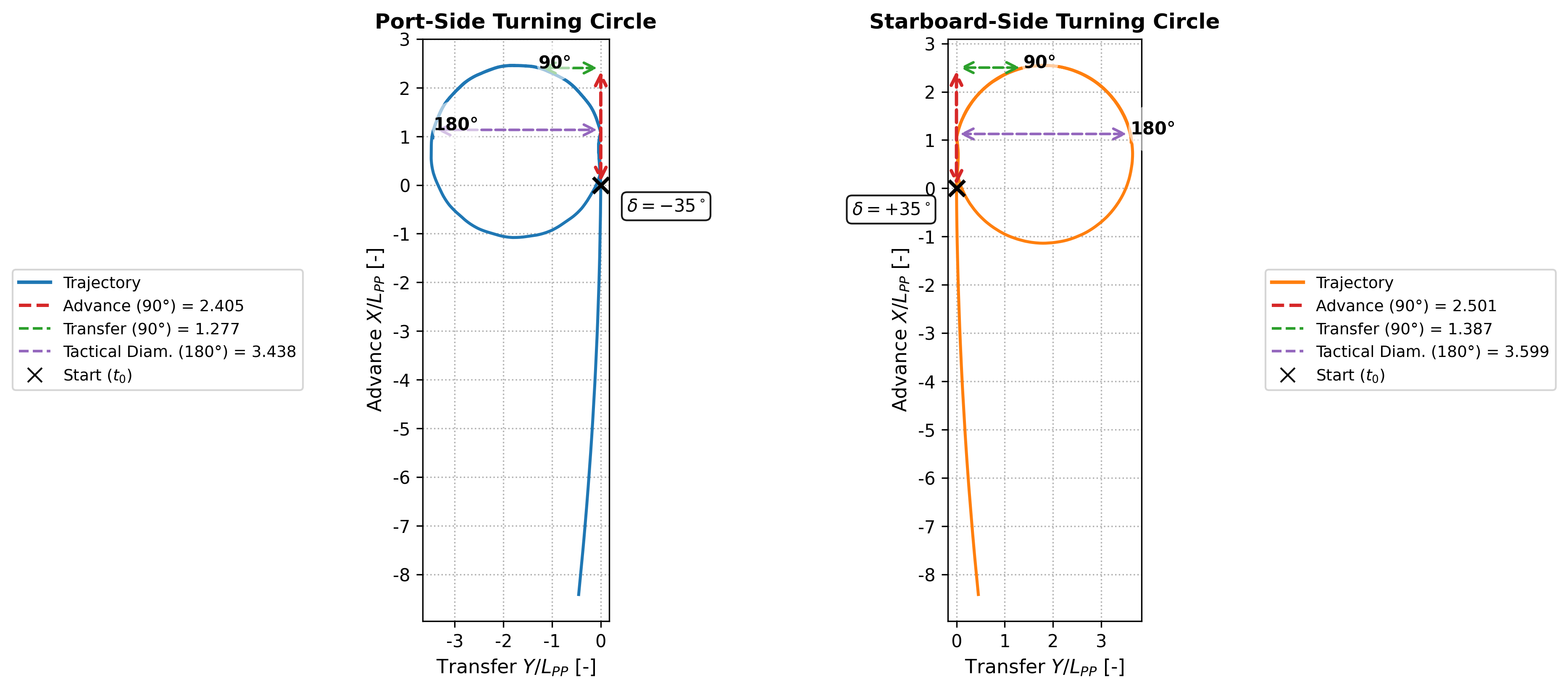}

    \caption{
    TC trajectories for port and starboard manoeuvres in normalised longitudinal and transverse coordinates 
    ($X/L_{PP}$, $Y/L_{PP}$). 
    Supplementary simulation videos are available for the corresponding manoeuvres:
    \href{https://www.youtube.com/watch?v=pQFr0AR-pdA}{Supplementary Video 1: TC Port} and
    \href{https://www.youtube.com/watch?v=2SdToAefDQU}{Supplementary Video 2: TC Starboard}.
    }
    \label{fig:tc_results}
\end{figure}

Figure~\ref{fig:tc_results} shows the TC trajectories for both port and starboard directions, with corresponding metrics summarised in Table~\ref{tab:tc_metrics}. The results show closely similar turning behaviour. Normalised advance ($Ad/L_{PP}$) differs by approximately $3.9\%$, while the tactical diameter ($Dt/L_{PP}$) differs by approximately $4.6$--$4.7\%$ between directions, indicating good directional symmetry while reflecting minor execution‑level asymmetries inherent in differential‑thrust actuation. All measured values satisfy the IMO MSC.137(76) \citep{IMO2002} criteria, with $Ad/L_{PP} \le 4.5$ and $Dt/L_{PP} \le 5.0$, confirming compliant steady‑turning behaviour and robust metric extraction.

\begin{table}[H]
\caption{TC manoeuvring metrics and IMO MSC.137(76) compliance.}
\label{tab:tc_metrics}
\centering
\footnotesize
\setlength{\tabcolsep}{3pt}
\renewcommand{\arraystretch}{1.15}

\begin{tabular}{lcccccc}
\toprule
\textbf{Manoeuvre} &
\makecell{\textbf{Advance}\\{$Ad/L_{PP}$}} &
\makecell{\textbf{Transfer}\\{$Tr/L_{PP}$}} &
\makecell{\textbf{Tactical}\\\textbf{Diam.}\\{$Dt/L_{PP}$}} &
\makecell{\textbf{Steady Turn}\\\textbf{Diam.}\\{$D/L_{PP}$}} &
\makecell{\textbf{IMO MSC.137}\\\textbf{Criteria}} &
\textbf{Result} \\
\midrule

TC to Port  
& 2.403
& 1.270
& 3.435
& 3.530
& $Ad/L_{PP}\!\le\!4.5,\;Dt/L_{PP}\!\le\!5.0$
& \textbf{PASS} \\

TC to Starboard  
& 2.497
& 1.369
& 3.597
& 3.686
& $Ad/L_{PP}\!\le\!4.5,\;Dt/L_{PP}\!\le\!5.0$
& \textbf{PASS} \\

\bottomrule
\end{tabular}
\end{table}

\subsection{Zig--Zag Manoeuvre Results}

\begin{figure}[H]
    \centering
    \includegraphics[width=\linewidth]{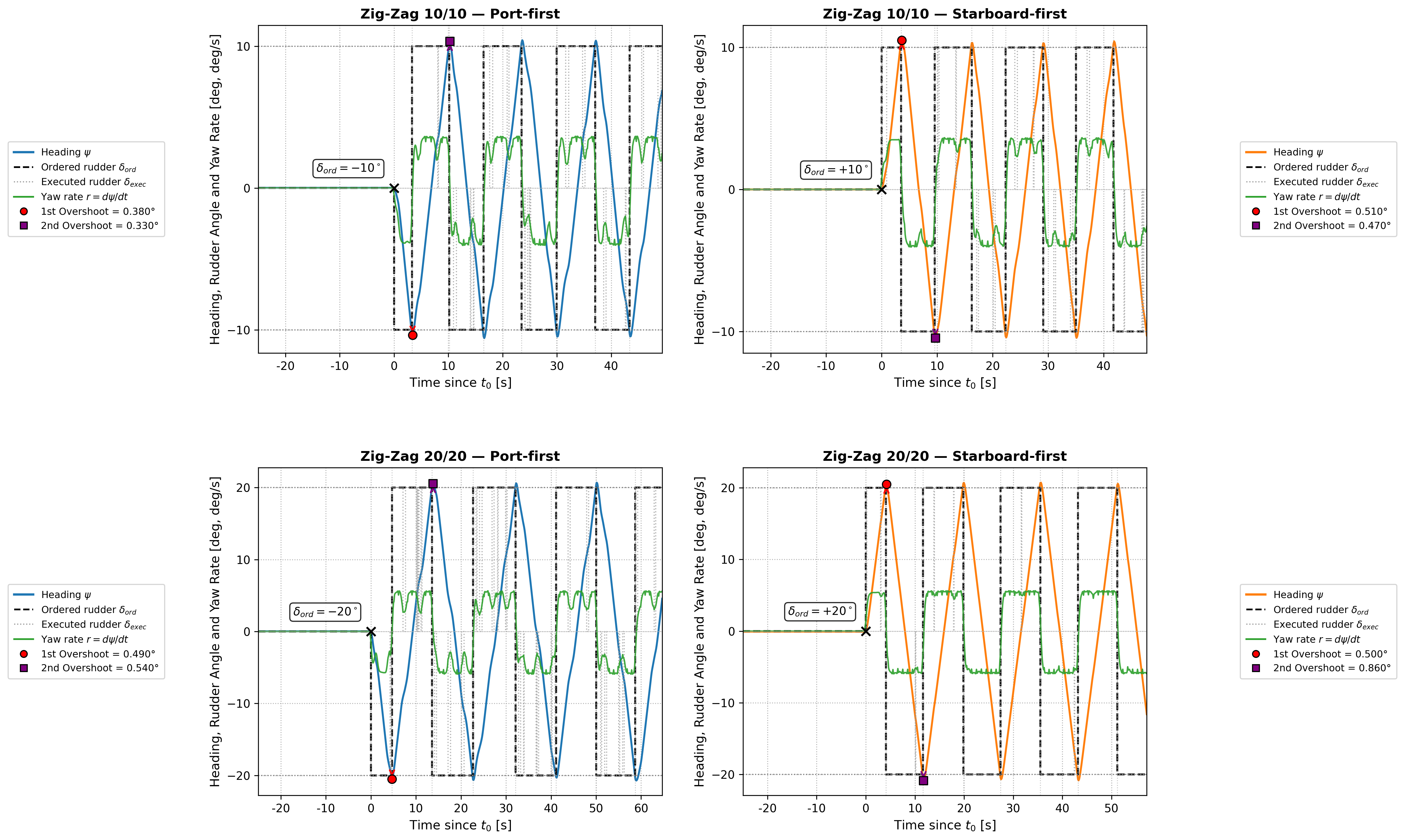}
    \caption{
    ZZ responses for $10^\circ/10^\circ$ and $20^\circ/20^\circ$ tests in port-first and starboard-first executions.
    The corresponding simulation recordings are provided as supplementary videos:
    \href{https://www.youtube.com/watch?v=sHBhFfHwwGk}{Video 3, ZZ $10^\circ/10^\circ$ Port-first},
    \href{https://www.youtube.com/watch?v=uo5LbihX1AM}{Video 4, ZZ $10^\circ/10^\circ$ Starboard-first},
    \href{https://www.youtube.com/watch?v=yCZDRFfJikc}{Video 5, ZZ $20^\circ/20^\circ$ Port-first}, and
    \href{https://www.youtube.com/watch?v=yEhqeSOkkQ4}{Video 6, ZZ $20^\circ/20^\circ$ Starboard-first}.
    }
    \label{fig:zz_results}
\end{figure}

Figure~\ref{fig:zz_results} shows representative ZZ responses for the $\pm10^\circ$ and $\pm20^\circ$ manoeuvres, including executed actuation, heading response, yaw rate, and overshoot angles.

\begin{table}[H]
\caption{ZZ manoeuvring metrics and IMO MSC.137(76) compliance.}
\label{tab:zz_metrics}
\centering
\footnotesize
\setlength{\tabcolsep}{3pt}
\renewcommand{\arraystretch}{1.15}

\begin{tabular}{lcccccc}
\toprule
\textbf{Manoeuvre} &
\makecell{\textbf{1st Heading}\\\textbf{Peak}\\{[deg]}} &
\makecell{\textbf{2nd Heading}\\\textbf{Peak}\\{[deg]}} &
\makecell{\textbf{Overshoot}\\\textbf{Excess}\\{[deg]}} &
\makecell{\textbf{Peak Yaw}\\\textbf{Rate}\\{[deg/s]}} &
\makecell{\textbf{IMO MSC.137}\\\textbf{Overshoot}\\\textbf{Criteria}} &
\textbf{Result} \\
\midrule

ZZ $10^\circ/10^\circ$ Port-first    
& 10.38 & 10.33 & 0.38 / 0.33 & 4.05
& $\le 10^\circ,\; \le 25^\circ$
& \textbf{PASS} \\

ZZ $10^\circ/10^\circ$ Starboard-first 
& 10.51 & 10.47 & 0.51 / 0.47 & 4.05
& $\le 10^\circ,\; \le 25^\circ$
& \textbf{PASS} \\

ZZ $20^\circ/20^\circ$ Port-first 
& 20.49 & 20.54 & 0.49 / 0.54 & 5.78
& $\le 25^\circ,\; \le 40^\circ$
& \textbf{PASS} \\

ZZ $20^\circ/20^\circ$ Starboard-first
& 20.50 & 20.38 & 0.50 / 0.38 & 5.78
& $\le 25^\circ,\; \le 40^\circ$
& \textbf{PASS} \\

\bottomrule
\end{tabular}
\end{table}

As summarised in Table~\ref{tab:zz_metrics}, the first and second overshoot excesses remain well below the IMO thresholds, with values below $1^\circ$ in all cases. The number of detected reversals is consistent across port-first and starboard-first tests, indicating reliable response-based switching. Additionally, peak yaw rates increase with manoeuvre magnitude, from approximately $4.05~\mathrm{deg/s}$ for $10^\circ/10^\circ$ tests to $5.78~\mathrm{deg/s}$ for $20^\circ/20^\circ$ tests, confirming compliant yaw-checking behaviour and consistent actuation-traceable execution.

\subsection{Actuation Traceability and Consistency Validation}
\begin{figure}[H]
    \centering
    \includegraphics[width=\linewidth, height=0.25\textheight, keepaspectratio]{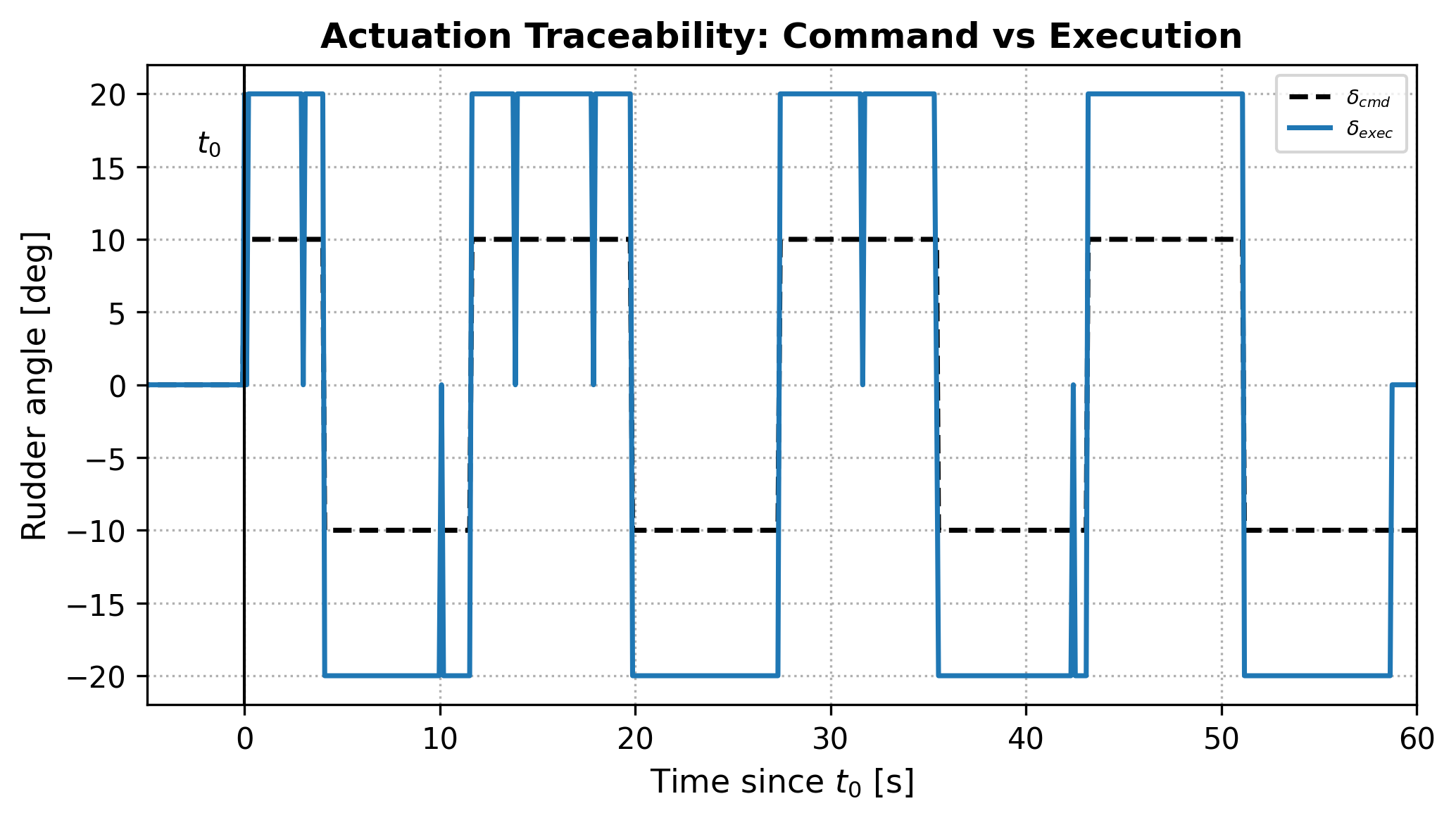}
    \caption{Command–execution traceability during ZZ $10^\circ/10^\circ$ Starboard-first manoeuvre.}
    \label{fig:rudder_traceability}
\end{figure}

Figure~\ref{fig:rudder_traceability} illustrates the relationship between the ordered rudder‑equivalent command $\delta_{\text{cmd}}$ and executed proxy $\delta_{\text{exec}}$ during a representative ZZ manoeuvre. 

The results demonstrate systematic deviations between commanded and realised actuation during saturation and reversal phases. Treating commanded inputs as directly realised forces would ignore these effects and introduce bias into manoeuvring metrics and subsequent hydrodynamic‑derivative estimation. The actuation‑traceability diagnostics therefore provide an essential validation layer, ensuring that SI is grounded in physically realised control actions.

\begin{figure}[H]
    \centering
    \includegraphics[width=\linewidth, height=0.25\textheight, keepaspectratio]{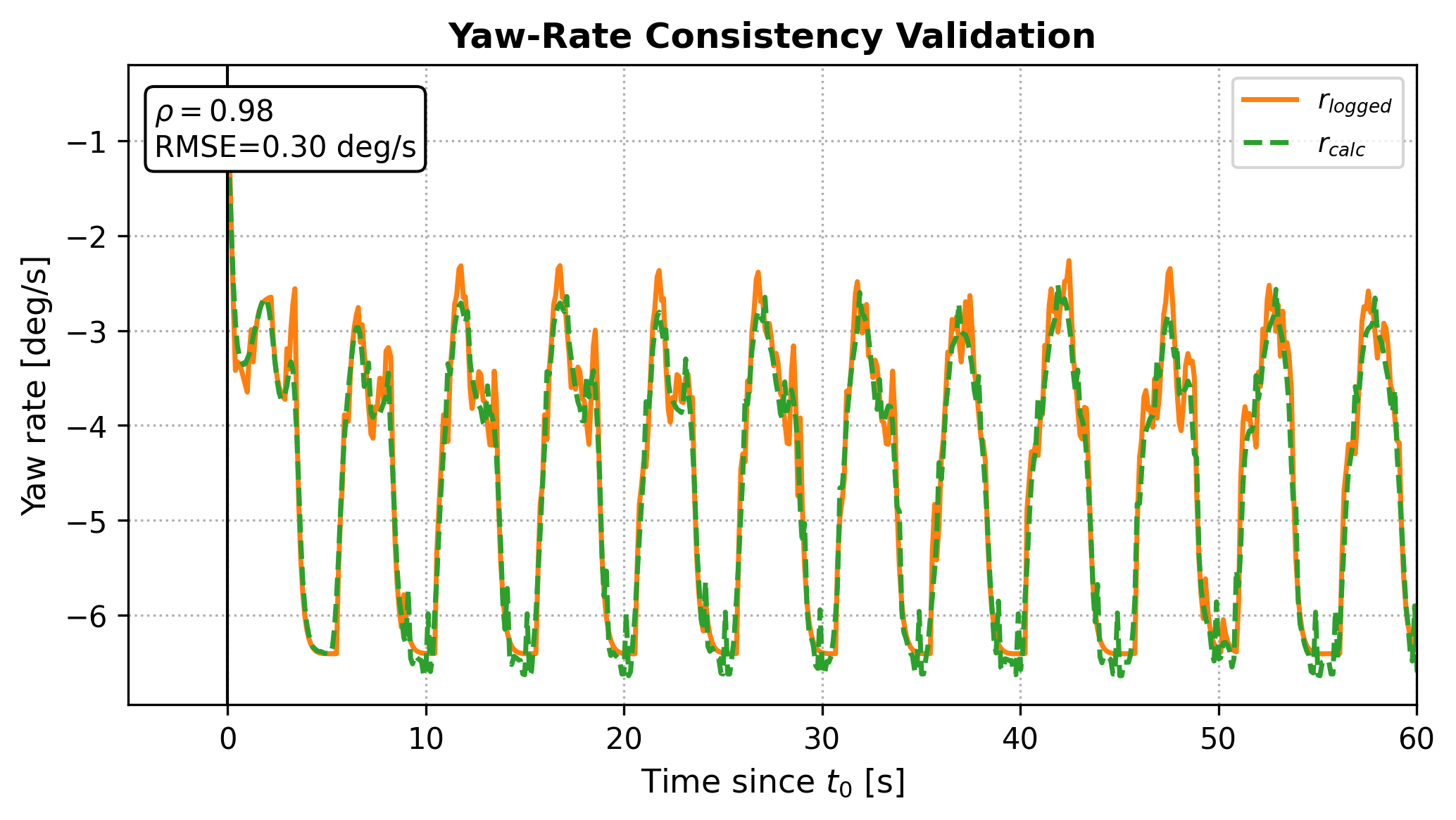}
    \caption{Yaw‑rate consistency validation comparing logged and reconstructed yaw rate for TC to Port.}
    \label{fig:yaw_validation}
\end{figure}

Figure~\ref{fig:yaw_validation}, on the other hand, compares the yaw rate logged directly from the Unity rigid‑body dynamics, $r_{\text{logged}}$, with the reconstructed yaw rate, $r_{\text{calc}}$. The two signals exhibit strong agreement, with high correlation ($\rho = 0.98$) and low root‑mean‑square error ($0.30~\mathrm{deg/s}$). Together with the manoeuvre‑level results, this provides system‑level confidence in the physical consistency and validity of the recorded datasets.

\section{Discussion}

The results demonstrate that manoeuvre behaviour in high‑fidelity marine simulators is governed by realised actuation rather than commanded inputs alone. Although symmetric rudder‑equivalent commands were issued, measurable execution‑level asymmetries were observed. For the TC manoeuvres, normalised advance differed by approximately 3.9\%, while tactical diameter differed by approximately 4.6--4.7\% between directions, despite identical commanded inputs. As evidenced by the separation between $\delta_{\text{cmd}}$ and $\delta_{\text{exec}}$, these differences arise from execution‑level effects such as actuator saturation and thrust imbalance asymmetry. 

For ZZ manoeuvres, overshoot excesses remained below $1^\circ$ in all cases, satisfying IMO MSC.137(76) limits for both $\pm10^\circ$ and $\pm20^\circ$ tests. However, peak yaw rates varied with manoeuvre magnitude and realised actuation, reaching approximately $4.1~\mathrm{deg/s}$ for $\pm10^\circ$ tests and $5.8~\mathrm{deg/s}$ for $\pm20^\circ$ tests.  Reporting peak yaw rate alongside overshoot metrics therefore captures both standards compliance and dynamic response intensity, providing a more complete assessment of yaw stability than overshoot angles alone. 
The actuation‑traceability analysis reveals that assuming perfect equivalence between commanded and realised actuation would hide physically relevant effects. The ratio $\eta = |\delta_{\text{exec}}|/|\delta_{\text{cmd}}|$ deviated from unity during saturation and reversal phases, directly influencing turning diameter, yaw‑rate peaks, and transient response timing. Without explicit command–execution separation, such effects would propagate into biased hydrodynamic‑derivative estimates during SI. The strong agreement between logged and reconstructed yaw rates (correlation $\rho = 0.98$, RMSE $0.30~\mathrm{deg/s}$) confirms internal physical consistency and validates the post‑processing pipeline. Overall, execution‑grounded virtual sea trials yield manoeuvring datasets that are not only IMO/ITTC‑compliant, but also richer and more informative for SI and DT calibration than command‑based simulator datasets.

\section{Conclusions and Future Work}

This paper presented an actuation-traceable virtual sea-trial framework for automated TC and ZZ manoeuvres in MARUS. By grounding timing, metric extraction, and validation in realised actuation, the framework generates IMO/ITTC-aligned datasets while preserving command--execution traceability. Results confirmed repeatable manoeuvre execution, compliant turning and yaw-checking behaviour, execution-level asymmetries, and strong signal consistency. Future work will introduce environmental disturbances, integrate the datasets into online SI, and validate transferability through physical USV trials for execution-aware DT development.

\section*{Acknowledgements}

This research was supported through Graduate Teaching Assistant (GTA) funding from Sheffield Hallam University’s School of Engineering and Built Environment. For the purpose of open access, the author has applied a Creative Commons Attribution (CC BY) licence to any Author Accepted Manuscript version arising from this submission. 
The dataset used in this study is available on Zenodo: Rezayan, P. (2026). iSCSS26 Vessel Manoeuvring Dataset: Traceable Virtual Sea Trials in the Marine Robotics Unity Simulator. Zenodo. https://doi.org/10.5281/zenodo.20627303, and the GitHub code repository can be accessed from the corresponding author upon request.

\bibliographystyle{elsarticle-harv}
\bibliography{Bibliography}










\end{document}